\documentclass[10pt,twocolumn,letterpaper]{article}

\usepackage{cvpr}              

\usepackage{graphicx}
\usepackage{amsmath}
\usepackage{amssymb}
\usepackage{booktabs}

\usepackage[table]{xcolor}
\usepackage{soul}
\usepackage{lipsum}
\usepackage[flushleft]{threeparttable}
\usepackage{multirow}

\usepackage[pagebackref,breaklinks,colorlinks]{hyperref}

\usepackage[capitalize]{cleveref}
\crefname{section}{Sec.}{Secs.}
\Crefname{section}{Section}{Sections}
\Crefname{table}{Table}{Tables}
\crefname{table}{Tab.}{Tabs.}

\DeclareMathOperator*{\argmin}{arg\,min}

\def\cvprPaperID{9076} 
\def\confName{CVPR}
\def\confYear{2022}

\graphicspath{{./figures/}}

\newcommand{\name} {TSC}

\newenvironment{Itemize}%
{
\setlength{\leftmargini}{9pt}%
\begin{itemize}%
\setlength{\itemsep}{0pt}%
\setlength{\topsep}{0pt}%
\setlength{\partopsep}{0pt}%
\setlength{\parskip}{0pt}}%
{\end{itemize}}

{
\setlength{\leftmargini}{9pt}%
\begin{enumerate}%
\setlength{\itemsep}{0pt}%
\setlength{\topsep}{0pt}%
\setlength{\partopsep}{0pt}%
\setlength{\parskip}{0pt}}%
{\end{enumerate}}

\newcommand{\red}[1]{\textcolor{black}{#1}}
\usepackage{comment}

\begin{document}

\title{Targeted Supervised Contrastive Learning for Long-Tailed Recognition}

\author{Tianhong Li$^{1,}$\thanks{Indicates equal contribution.} \quad Peng Cao$^{1,{\ast}}$ \quad Yuan Yuan$^1$ \quad Lijie Fan$^1$ \quad Yuzhe Yang$^1$ \\ 
\quad Rogerio Feris$^2$ \quad Piotr Indyk$^1$ \quad Dina Katabi$^1$ \\\\ $^1$MIT CSAIL, $^2$MIT-IBM Watson AI Lab}
\maketitle

\begin{abstract}

Real-world data often exhibits long tail distributions with heavy class imbalance, where the majority classes can dominate the training process and alter the decision boundaries of the minority classes. Recently, researchers have investigated the potential of supervised contrastive learning for long-tailed recognition, and demonstrated that it provides a strong performance gain. In this paper, we show that while supervised contrastive learning can help improve performance, past baselines suffer from poor uniformity brought in by imbalanced data distribution. This poor uniformity manifests in samples from the minority class having poor separability in the feature space. To address this problem, we propose targeted supervised contrastive learning (\name), which improves the uniformity of the feature distribution on the hypersphere. \name\ first generates a set of targets uniformly distributed on a hypersphere. It then makes the features of different classes converge to these distinct and uniformly distributed targets during training.  This forces all classes, including minority classes, to maintain a uniform distribution in the feature space, improves class boundaries, and provides better generalization even in the presence of long-tail data. Experiments on multiple datasets show that \name~achieves state-of-the-art performance on long-tailed recognition tasks. Code is available \href{https://github.com/LTH14/targeted-supcon}{here}.
\end{abstract}
 
\section{Introduction}\label{sec:intro}

Real-world data often has a long tail distribution over classes: A few classes contain many instances (head classes), whereas most classes contain only a few instances (tail classes). For critical applications, such as medical diagnosis, autonomous driving, and fairness, the data are by their nature heavily imbalanced, and the minority classes are particularly important (minority classes can be patients or accidents~\cite{shen2015long,yang2022multi,yang2021delving}). Interest in such problems has motivated much recent research on imbalanced classification, where the training dataset is imbalanced or long-tailed but the test dataset is equally distributed among classes~\cite{kang2019decoupling,wang2020long,yang2020rethinking,cao2019learning,yang2022multi}. 

Long-tailed and imbalanced datasets pose major challenges for classification tasks leading to a significant performance drop~\cite{ando2017deep, buda2018systematic, collobert2008unified, yang2019me, wu2020solving}. Techniques such as data re-sampling~\cite{chawla2002smote, shen2016relay, buda2018systematic, ando2017deep} and  loss re-weighting~\cite{cao2019learning, cui2019class, dong2018imbalanced, khan2019striking, khan2017cost, byrd2019effect} can improve the performance of tail classes but typically harm head classes \cite{kang2019decoupling}.
Recently, researchers have investigated the potential of supervised contrastive learning for long-tailed recognition, and demonstrated that it provides a strong performance gain~\cite{kang2020exploring}. They further proposed $k$-positive contrastive learning (KCL), a variant of supervised contrastive learning that yields even better performance on long-tailed datasets.

\begin{figure}[t]
\begin{center}
\includegraphics[width=0.48\textwidth]{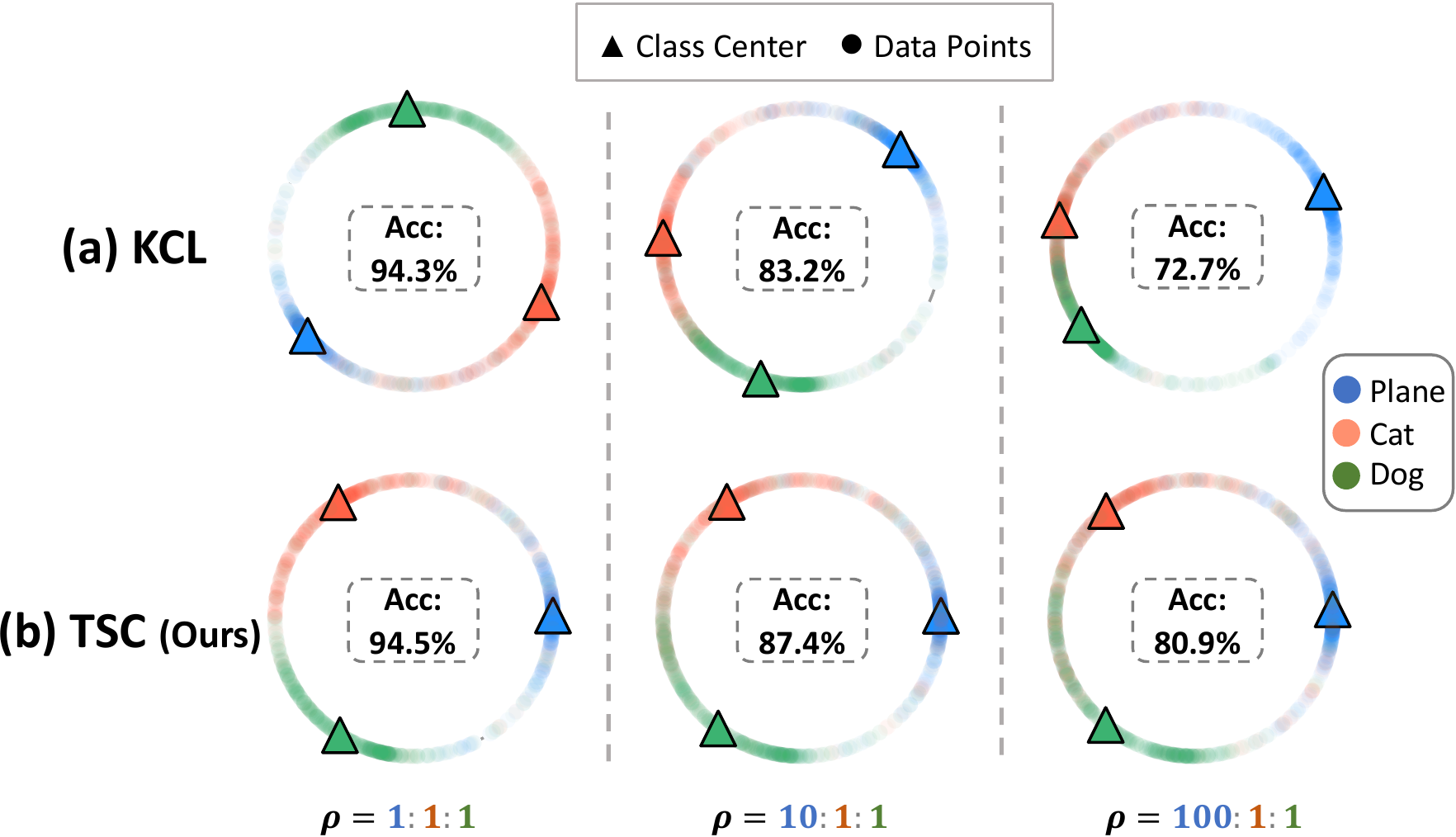} 
\end{center}
\vspace{-10pt}
\caption{\small
Test data feature distribution of (a) k-positive contrastive learning (KCL) and (b) \name\ for three classes of CIFAR10 (plane, cat, dog), for different training data imbalance ratios $\rho$. With high imbalance ratio, class centers learned by KCL exhibit poor uniformity while class centers learned by \name~are still uniformly distributed and thus \name\ achieves better performance (where Acc refers to Accuracy on test data).}
\label{fig:intro}
\vspace{-10pt}
\end{figure}

However, while supervised contrastive learning can be beneficial, applying the contrastive loss (including the KCL loss) to imbalanced data can yield poor uniformity, which hampers performance. Uniformity is a desirable property~\cite{wang2020long}; it refers to that in an ideal scenario supervised contrastive learning should converge to an embedding where the different classes are uniformly distributed on a hypersphere~\cite{wang2020long,graf2021dissecting}. Uniformity maximizes the distance between classes in the feature space, i.e., maximizes the margin. As a result, it improves  generalizability. 

But, when the classes are imbalanced, training naturally puts more weight on the loss of majority classes and less weight on that of minority classes. As a result, the classes are no longer uniformly distributed in the feature space. To illustrate this issue, we consider three classes from CIFAR-10: dog, cat, and plane. We train a KCL model~\cite{kang2020exploring} on this data for different imbalance ratios, $\rho$. For visualization clarity we use a 2D feature space. As seen in Fig.~\ref{fig:intro}(a), when the classes are balanced (i.e., $\rho$=1:1:1), the centers of the three classes are uniformly distributed in the KCL feature space. In contrast, when the imbalance ratio is high  (e.g., $\rho$=100:1:1), the classes with fewer training instances start to collapse into each other, leading to unclear and inseparable decision boundaries, and thus lower performance. This is because the imbalanced data distribution naturally puts more weight on the uniformity loss between the head class and the tail classes, and less weight on that between the two tail classes, making the distance between head and tail classes much larger than the distance between two tail classes. The more imbalanced the long-tailed data, the more biased and less uniformly distributed the feature space.

One may attempt to fix this problem by oversampling the tail classes or re-weighting the loss function. However,  as shown in~\cite{kang2019decoupling}, those methods overfit tail classes and improve tail-class performance at the expense of head classes, and thus harm the quality of the learned features. 
Therefore, a method that performs instance-balanced sampling while still being able to learn a uniform feature space is needed. 

In this paper, we propose targeted supervised contrastive learning (\name) for long-tailed recognition. To avoid the feature space being dominated and biased by head classes, we generate the optimal locations of class centers in advance (i.e., off-line). We call these uniformly distributed points class targets. We then devise 
 an online matching-training scheme that performs contrastive training while adaptively matching samples from each class to one of the targets. As shown in Fig. \ref{fig:intro}(b), \name~learns a class-balanced feature space regardless of the imbalance ratio of the training set.
 
Note that one cannot simply match any target point with any class. Though the targets are uniformly distributed in the feature space, the distance between two targets can vary widely. For example, if the number of classes in Fig. \ref{fig:intro} was 10 instead of 3, then though the targets are uniformly distributed, some targets will be closer to each other than the rest. Thus, our matching-training scheme has to ensure that classes that are semantically close (e.g., cat and dog)
converge to nearby targets, and classes that are semantically farther apart converge to relatively distant targets. 

We evaluate \name~on long-tailed benchmark datasets including CIFAR-10-LT, CIFAR-100-LT, ImageNet-LT, and iNaturalist, and show that it improves the state-of-the-art (SOTA) performances on all of them.
 
To summarize, this paper makes the following contributions:
\begin{Itemize}
    \item It introduces \name, a novel framework for long-tailed recognition that avoids the feature space being dominated and biased by head classes.
    \item It empirically shows that supervised contrastive learning baselines can suffer from poor uniformity when applied to long-tailed recognition, which degrades their performances.
    \item It further shows that \name~achieves SOTA long-tailed recognition performances on benchmark datasets, demonstrating the effectiveness of the proposed method.
\end{Itemize}

\section{Related Works}

\textbf{Imbalanced Learning and Long-tailed Recognition.}
Real-world data typically follows a long-tailed or imbalanced distribution, which biases the learning towards head classes, and degrades performance on tail classes~\cite{yang2022multi, zhang2021deep}. Conventional methods have focused on designing class re-balancing paradigms through data re-sampling~\cite{chawla2002smote, shen2016relay, buda2018systematic, ando2017deep} or adjusting the loss weights for different classes during training~\cite{cao2019learning, cui2019class, dong2018imbalanced, khan2019striking, khan2017cost, byrd2019effect}. However, such methods improve tail class performance at the expense of head class performance~\cite{kang2019decoupling}.
Researchers have also tried to improve long-tailed recognition by using ensembles over different data distributions\cite{zhou2020bbn, wang2020long,zhang2021test}, i.e., they re-organize long-tailed data into groups, train a model per group, and combine individual models in a multi-expert framework, or by using a distillation label generation module guided by self-supervision\cite{li2021self}. Ensemble-based and distillation-based methods have been shown to be orthogonal to methods operating on a single model and could leverage improvements in single-model methods to improve the performance.

Recent works~\cite{kang2019decoupling,zhou2020bbn,yang2021delving} show that decoupling representation learning from classifier learning can lead to good features, which motivates the use of feature extractor pre-training for long-tailed recognition. The authors of~\cite{yang2020rethinking} introduced a self-supervised pre-training initialization that alleviates the bias caused by imbalanced data. The authors of~\cite{kang2020exploring} further show that self-supervised learning can improve robustness to data imbalance, and introduce k-positive contrastive learning (KCL). 
Our work builds on this literature, and introduces a new approach to dealing with data imbalance using pre-computed uniformly-distributed targets that guide the training process to achieve better uniformity and improved class boundaries.

\textbf{Contrastive Learning.}
Recent years have witnessed a steady progress on self-supervised representation learning~\cite{oord2018representation, doersch2015unsupervised,huang2017multi, noroozi2016unsupervised,fan2018end,fan2019controllable,fan2017adversarial,li2021self, fan2020learning,li2019making,fan2020home,zhao2018through,zhao2018rf}. Contrastive learning~\cite{he2020momentum,chen2020simple, chen2021exploring, khosla2020supervised,fan2021does, wang2020understanding, li2020making, grill2020bootstrap} has been outstandingly successful on multiple tasks~\cite{li2022rf, wu2021consistency, morgado2021robust}. The core idea of contrastive learning is to align the positive sample pairs and repulse the negative sample pairs. Many works~\cite{wang2020understanding, tian2020makes, li2020making, tian2021understanding, graf2021dissecting} have made efforts to understand and explain its properties, as well as its effects on downstream tasks. 
In particular, \cite{wang2020understanding} proves that contrastive learning asymptotically optimizes both
alignment (closeness) of features from positive pairs, and uniformity of the induced distribution of the (normalized) features on the hypersphere. 
 Supervised contrastive learning (SupCon)~\cite{khosla2020supervised} extends contrastive learning to the fully-supervised setting. It selects the positive samples from data belonging to the same class and aligns them in the embedding space, while simultaneously pushing away samples from different classes. By leveraging class labels with a contrastive loss, SupCon surpasses the performance  of the traditional supervised cross-entropy loss on image classification. Most work on contrastive learning focuses on balanced data. Recently however, researchers have applied contrastive learning to imbalanced and long-tailed classification and demonstrated improved performance\cite{kang2020exploring,yang2020rethinking}.

\section{Method}

\name~is a training framework for improving the uniformity of the latent feature distribution. It aims to learn representations where the centers of each class are distributed uniformly on a hypersphere, and thus obtain clear decision boundaries between classes. \name~is especially effective on long-tailed recognition tasks, since for traditional methods based on supervised contrastive loss, classes with fewer training instances could easily collapse with other tail classes, resulting in poor classification performance.

\begin{figure}[t]
\begin{center}

\includegraphics[width=0.46\textwidth]{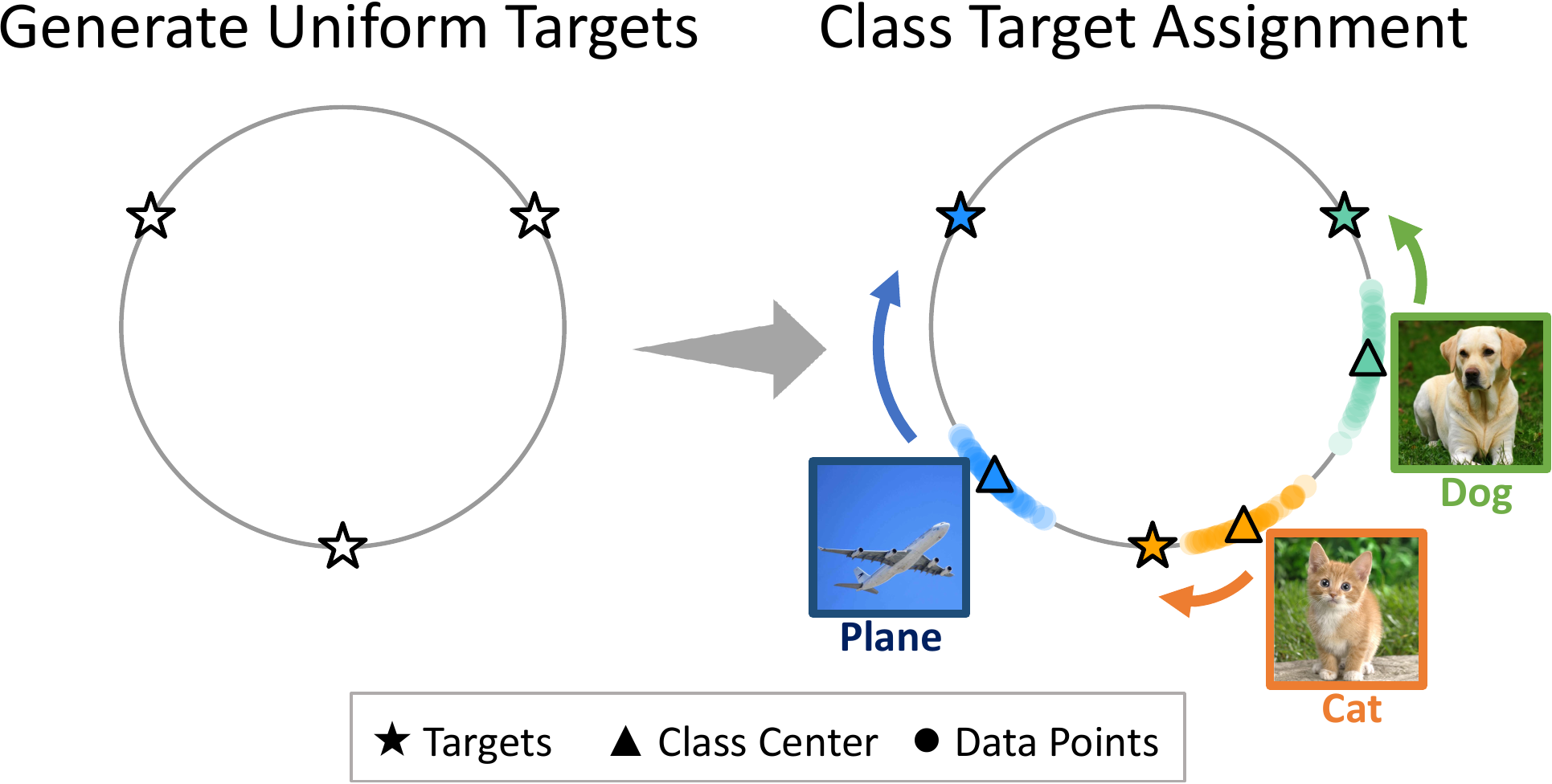}
\end{center}
\vspace{-10pt}
\caption{\small Illustration of \name. It first computes the optimal targets for the class centers on the hypersphere. Then, during training, in each iteration, each target is assigned to the nearest class, and a targeted contrastive learning loss is designed to encourage the samples from each class to move to the assigned target position. }
\label{fig:model}
\vspace{-10pt}
\end{figure}

Fig. \ref{fig:model} shows the overview of \name. The targets where we want to position the class centers on the hypersphere are pre-computed prior to training and kept fixed thereafter. During training, the target positions are assigned to the classes online, and a targeted supervised contrastive loss is designed to encourage samples from each class to move to the assigned target position.

\subsection{Target Generation}

We first compute the optimal positions for the targets in the feature space. Since the test dataset is equally distributed among classes in long-tailed recognition, and the features in contrastive learning are positioned on a unit-hypersphere $\mathcal{S}^{d-1} = \{u \in \mathbb{R}^d: ||u|| = 1\}$~\cite{chen2020simple,he2020momentum}, 
the ideal class targets should be uniformly distributed on this hypersphere. 
Note that computing these ideal target positions does not require access to the data, and only requires knowing the number of classes and the dimension of the feature space. 
Thus, similar to the uniformity loss defined in \cite{wang2020understanding}, we design the target positions of $C$ classes, $\{t^*_i\}_{i=1}^C$, as the minimizer of 
\begin{equation}
    \mathcal{L}_{u}(\{t_i\}_{i=1}^C) = \frac{1}{C}\sum\limits_{i=1}^C \log \sum\limits_{j=1}^C e^{t_i^T\cdot t_j/\tau}.
\end{equation}



Intuitively, we want the target positions on the hypersphere to be as far away from each other as possible. The ideal locations would be perfectly uniformly distributed on the hypersphere, forming the vertices of a regular simplex (i.e. $\exists \delta \in \mathbb{R}, t_i^T\cdot t_j = \delta, \forall i,j, i\neq j$ and $\sum\limits_i t_i = 0$)\cite{graf2021dissecting}. 
However, when the dimension of the hypersphere is not large enough (e.g., $d<(C-1)$), computing the minimum of the above equation analytically becomes very hard~\cite{graf2021dissecting}.\footnote{Even for the well-studied Thomson problem, which tries to determine the most uniform (minimum electrostatic potential energy) arrangement of $N$ electrons on a 3D sphere, the solutions are known only for $N = \{1,2,3,4,5,6,12\}$.} Therefore, $\{t^*_i\}_{i=1}^C$ are calculated by gradient descent on $\mathcal{L}_{u}$, where $t_i$ is restricted to be on the hypersphere. Note that the minimum of $\mathcal{L}_{u}$ after gradient descent will be equal to its analytical minimum when $d\geq(C-1)$.



\subsection{Matching-Training Scheme}

\noindent\textbf{Class-Target Assignment.} Once we obtain a set of target positions, we need to assign a class label to each target. One way to do so is to randomly assign class labels to target positions. However, this will lead to a feature space with very poor semantics. This is because some target positions may be close to each other on the hypersphere and some are far away, especially when the number of classes is large (e.g., ImageNet and iNaturalist). Ideally, classes that are semantically close to each other should be assigned to target positions that are also close to each other.

It is hard however to accurately quantify semantic closeness between two classes. Even if we could quantify it, i.e., there is a well-defined ``semantic distance'' between two classes, it is computationally hard (i.e., no poly-time solution)
to compute the optimal assignment that matches classes with target positions while keeping the semantic distance between classes consistent with the euclidean distance between their targets.\footnote{This problem can be formulated as the matching problem between two adjacency matrices. The graph isomorphism problem \cite{fortin1996graph}, which is not known to be solvable in polynomial time yet, can be reduced to this problem.} 

To solve this problem, we design a heuristic algorithm that finds a good assignment, while preserving the semantic structure of the feature space. Instead of pre-computing the assignment, we do it adaptively during training. Specifically,  
after each iteration in the training process, we use the Hungarian Algorithm \cite{kuhn1955hungarian} to find the assignment $\{\sigma^*_i\}_{i=1}^C$  that minimizes the distance between the target positions and the normalized class centers assigned to them, i.e.:
\begin{equation}
    \{\sigma^*_i\}_i =  \argmin\limits_{\{\sigma_i\}_i} \frac{1}{C}\sum\limits_{i=1}^C ||t_{\sigma_i} - c_i||,
\end{equation}
where $c_i = \frac{\sum\limits_{v_j \in F_i} v_j}{||\sum\limits_{v_j \in F_i} v_j||_2}$ and $F_i$ is the set of features from class $i$. In practice, since a batch may contain only a subset of all classes, we keep track of the centers of each class using a weighted moving average. To be specific, in every iteration, we compute the new class centers $c_i'$ for each class in the batch, and update the recorded $c_i$ by $c_i \gets 0.9\cdot c_i + 0.1\cdot c_i'$. As shown in Sec. \ref{subsec:ablation}, this  assignment algorithm demonstrates good performance on preserving the semantic structure of the feature space.



\noindent\textbf{Targeted Supervised Contrastive Loss.} To leverage the assigned targets, we design a targeted supervised contrastive loss $\mathcal{L}_{TSC}$. Given a batch of data samples $\{(x_i, y_i)\}_{i=1}^N$, where $y_i \in [C]$ is the class label of $x_i$. Define $v_i$ as the features of $x_i$ on the unit-hypersphere $\mathcal{S}^{d-1}$, $\Tilde{v_i}$ as the features generated by augmenting $x_i$, $V_i =  \{v_n\}_{n=1}^N \backslash \{v_i\} $ as the current batch of features excluding $v_i$ and $V_{i, k}^+ \subseteq V_i$ the positive set of $v_i$ containing $k$ features uniformly drawn from $\{v_j \in V_i: y_j = y_i\}$. Let $\Tilde{V_i} = \{\tilde{v_i}\}\cup V_i$ and $\Tilde{V_{i, k}^+} = \{\Tilde{v_i}\}\cup V_{i, k}^+$. $\mathcal{L}_{TSC}$ is then defined as:

\begin{gather}
\begin{aligned}
     \mathcal{L}_{TSC} = -\frac{1}{N} \sum_{i=1}^N \Big(&\frac{1}{k+1}\sum_{v_j^+ \in \Tilde{V^+_{i, k}}}  \log \frac{e^{v_i^T\cdot v_j^+/\tau}}{ \sum\limits_{v_j\in \Tilde{V_i} \cup U} e^{v_i^T\cdot v_j/\tau}}\\
    &+ \lambda  \log \frac{e^{v_i^T\cdot c^*_i/\tau}}{ \sum\limits_{v_j\in \Tilde{V_i} \cup U} e^{v_i^T\cdot v_j/\tau}}\Big),
\end{aligned}
\label{eq:loss}
\raisetag{30pt}
\end{gather}
where $U=\{t^*_i\}_{i=1}^C$ is the set of pre-compute targets, and $c^*_i = t^*_{\sigma^*_{y_i}}$ is the assigned target of the corresponding class. Note that the loss is the sum of two components. The first is a standard contrastive loss as used by KCL~\cite{kang2020exploring}, whereas the second is a contrastive loss between the target and the samples in the batch. This latter loss moves the samples closer to the target of their class and away from the targets of other classes.

$\mathcal{L}_{TSC}$ forces projections from each class to be aligned with its assigned target, while distributing the targets uniformly on the hypersphere, and thus is  beneficial for long-tailed recognition tasks. 



\section{Experiments}
\label{sec:results}

We evaluate \name~on multiple long-tailed benchmark datasets and demonstrate its superior performances.

\subsection{Experiment Setup}
\label{subsec:setup}
We perform extensive experiments on benchmark datasets, like CIFAR-10-LT and CIFAR100-LT (The MIT License), and large-scale long-tailed datasets, such as ImageNet-LT (CC BY 2.0 license) \cite{liu2019large} and iNaturalist (CC0, CC BY or CC BY-NC license) \cite{van2018inaturalist}.  CIFAR10-LT and CIFAR-100-LT are sampled with an exponential decay across classes. The imbalance ratio $\rho$ is defined as the number of samples in the most frequent class divided by that of the least frequent class. Similar to previous works, we evaluate \name~on imbalance ratios of 10, 50 and 100 and a ResNet-32 backbone \cite{yang2020rethinking}. For ImageNet-LT and iNaturalist, we evaluate \name\ on a ResNet-50 backbone \cite{kang2020exploring,kang2019decoupling,wang2020long}.

Following \cite{kang2019decoupling, kang2020exploring}, we implement \name~on long-tailed recognition datasets using a two-stage training strategy. In the first stage, we train the representation encoder with the \name~loss. In the second stage, we train a linear classifier on top of the learned representation. For CIFAR-10-LT and CIFAR-100-LT, the linear classifier is trained with LDAM loss and class re-weighting. For ImageNet-LT and iNaturalist, the linear classifier is trained with CE loss and class-balanced sampling. We also empirically find that in the early training, it is better to first warm up the network by not assigning targets and training the network with just the KCL loss. Therefore, for ImageNet-LT and iNaturalist, we start the class target assignment after half of the total epochs. Following the KCL loss, we use $k=6$ for the TSC loss. We use the same data augmentations as previous works, including the non-contrastive learning baselines. We provide a detailed description of the implementation of our method in the appendix. All results are averaged over 3 trials with different random seeds.

There are mainly two types of works on long-tailed recognition: 1) single model training scheme design, such as new sampling strategies \cite{kang2019decoupling} or new losses \cite{cui2019class, cao2019learning, kang2020exploring}, and 2) ensembling over different data distributions, which re-organizes long-tailed data into groups, trains a model per group, and combines individual models in a multi-expert framework. Prior work~\cite{wang2020long} shows that these two approaches are orthogonal and can be combined together to improve performance. Our work falls in the first type of works. Therefore, we first compare \name~with  established state-of-the-art single model baselines, including \cite{cao2019learning, kang2019decoupling, kang2020exploring, kim2020m2m}, and then show that the combination of \name~and ensemble-based models can further improve its performance. Also, the literature compares different baselines for different datasets~\cite{yang2020rethinking,kang2020exploring,wang2020long}. \red{Thus, for each dataset, we compare with the typical and SOTA baselines for that dataset. }

\subsection{Results}

\textbf{CIFAR-10-LT \& CIFAR-100-LT.} Table \ref{tab:cifar} compares \name~with state-of-the-art baselines on CIFAR-10-LT and CIFAR-100-LT. It shows that, unlike other SOTA methods, \name~demonstrates consistent improvements over all baselines on all imbalance ratios in both datasets. This demonstrates that \name~can be generalized to different imbalance ratios and datasets easily as its design does not require prior knowledge of the imbalance ratio of the dataset.

\begin{table}[t]
\caption{\small Top-1 accuracy (\%) of ResNet-32 on long-tailed CIFAR-10 and CIFAR-100. \name\ consistently improves on past imbalanced learning techniques and achieves the best performance. Previous SOTA results for each imbalance ratio are colored with gray. We report the accuracy of our re-implemented KCL ($\dagger$) since they do not report their performance on CIFAR in \cite{kang2020exploring}.}
\vspace{-5mm}
\label{tab:cifar}
\small
\begin{center}
\resizebox{.48\textwidth}{!}{
\begin{tabular}{c|c|c|c|c|c|c}
\toprule
Dataset    &   \multicolumn{3}{c|}{CIFAR-10-LT}    &    \multicolumn{3}{c}{CIFAR-100-LT}   \\
\midrule
Imbalance Ratio ($\rho$)  & 100    &    50    &    10    &   100    &   50   &   10   \\
\midrule\midrule
CE               & 70.4 & 74.8 & 86.4 & 38.3 & 43.9 & 55.7  \\
CB-CE~\cite{cui2019class}    & 72.4 & 78.1 & 86.8 & 38.6 & 44.6 & 57.1  \\
Focal~\cite{lin2017focal} & 70.4 & 76.7 & 86.7 & 38.4 & 44.3 & 55.8 \\
CB-Focal~\cite{cui2019class} & 74.6 & 79.3 & 87.1 & 39.6 & 45.2 & 58.0  \\
CE-DRW~\cite{cao2019learning} & 75.1 & 78.9 & 86.4 & 40.5 & 44.7 & 56.2  \\
CE-DRS~\cite{cao2019learning} & 74.5 & 78.6 & 86.3 & 40.4 & 44.5 & 56.1  \\
LDAM~\cite{cao2019learning} & 73.4 & 76.8 & 87.0 & 39.6 & 45.0 & 56.9  \\
LDAM-DRW~\cite{cao2019learning} & 77.0 & 80.9 & \cellcolor{gray!25}88.2 & 42.0 & 46.2 & \cellcolor{gray!25}58.7   \\
M2m-ERM~\cite{kim2020m2m} & 78.3 & - & 87.9 & 42.9 & - & 58.2 \\
M2m-LDAM~\cite{kim2020m2m} & \cellcolor{gray!25}79.1 & - & 87.5 & \cellcolor{gray!25}43.5 & - & 57.6 \\
KCL$\dagger$ \cite{kang2020exploring} & 77.6 & \cellcolor{gray!25} 81.7 & 88.0 & 42.8 & \cellcolor{gray!25} 46.3 & 57.6  \\
\midrule
\name & \textbf{79.7} & \textbf{82.9} & \textbf{88.7}         & \textbf{43.8}  &  \textbf{47.4}  & \textbf{59.0} \\
\bottomrule
\end{tabular}
}
\end{center}
\vspace{-15pt}
\end{table}

\textbf{ImageNet-LT.} Table \ref{tab:imagenet} compares \name~with state-of-the-art baselines on ImageNet-LT dataset. As shown in the table, \name~demonstrates significant improvements over the baselines based on cross-entropy loss. It outperforms $\tau$-norm by 5.7\%, cRT by 5.1\%, and LWS by 4.7\%. It also improves over KCL \cite{kang2020exploring} for all class splits (1.1\% for many, 0.7\% for medium and 0.9\% for few). Note that \name~improves both the accuracy of the many split and that of the few split. This is because \name~not only improves the uniformity of the minority classes, but also improves the overall uniformity of the whole feature space, as is detailed in Sec. \ref{subsec:metrics}. This further demonstrates the effectiveness of the proposed \name~loss in improving uniformity across class centers and delivering clean boundaries between classes.

\begin{table}[t]
\caption{\name~outperforms previous state-of-the-art single-model methods on ImageNet-LT. Previous SOTA results of each class split (many, medium, few, all) are colored with gray. \red{Please note that the KCL accuracy for each class split reported in \cite{kang2020exploring} does not match the reported accuracy on all classes (61.8*0.385+49.4*0.479+30.9*0.136=51.658 which cannot be rounded to 51.5), indicating that their reported results may have a typo. Therefore, we also report the result of our re-implemented KCL (denoted with $\dagger$), which achieves similar accuracy on all classes but slightly different accuracy on each split.}}
\vspace{-5mm}
\label{tab:imagenet}
\begin{center}
\resizebox{.41\textwidth}{!}{
\begin{tabular}{c|cccc}
\toprule
Methods  & Many & Medium & Few & All \\ 
\midrule
OLTR \cite{liu2019large} & 35.8 & 32.3 & 21.5 & 32.2 \\
$\tau$-norm \cite{kang2019decoupling} & 56.6 & 44.2 & 27.4 & 46.7 \\ 
cRT \cite{kang2019decoupling} & 58.8 & 44.0 & 26.1 & 47.3 \\
LWS \cite{kang2019decoupling} & 57.1 & 45.2 & 29.3 & 47.7 \\
FCL \cite{kang2020exploring} & 61.4 & 47.0 & 28.2 & 49.8 \\
KCL \cite{kang2020exploring} & \cellcolor{gray!25}61.8 & \cellcolor{gray!25}49.4 & \cellcolor{gray!25}30.9 & \cellcolor{gray!25}51.5 \\
KCL $\dagger$ & 62.4 & 49.0 & 29.5 & 51.5 \\
\midrule
\name & \textbf{63.5} & \textbf{49.7} & \textbf{30.4} & \textbf{52.4} \\
\bottomrule
\end{tabular}
}
\end{center}
\vspace{-5pt}
\end{table}

\textbf{iNaturalist.} Table \ref{tab:inaturalist} compares \name~with state-of-the-art baselines on iNaturalist dataset. \name\ achieves the best performance among all baselines on all class splits, demonstrating its effectiveness in solving real-world long-tailed recognition problems such as natural species classification.

\begin{table}[t]
\vspace{-6pt}
\caption{\name~outperforms previous state-of-the-art single-model methods on challenging \textbf{iNaturalist 2018} \cite{van2018inaturalist} dataset, which contains 8142 classes. Previous SOTA results for each class split (many, medium, few, all) are colored with gray.}
\vspace{-5mm}
\label{tab:inaturalist}
\begin{center}
\resizebox{.45\textwidth}{!}{
\begin{tabular}{c|cccc}
\toprule
Methods  & Many & Medium & Few & All  \\ 
\midrule
CE & 72.2 & 63.0 & 57.2 & 61.7 \\ 
\midrule
CB-Focal  & - & - & - & 61.1 \\ 
OLTR \cite{liu2019large} & 59.0 & 64.1 & 64.9 & 63.9  \\ 
LDAM + DRW \cite{cao2019learning}  & - & - & - & 64.6 \\ 
cRT \cite{kang2019decoupling}  & \cellcolor{gray!25}69.0 & 66.0 & 63.2 & 65.2 \\
${\tau}$-norm \cite{kang2019decoupling} & 65.6 & 65.3 & \cellcolor{gray!25}65.9 & 65.6 \\
LWS \cite{kang2019decoupling} & 65.0 & \cellcolor{gray!25}66.3 & 65.5 & 65.9 \\
KCL \cite{kang2020exploring} & - & - & - & \cellcolor{gray!25}68.6 \\
\midrule
\name & \textbf{72.6} & \textbf{70.6} & \textbf{67.8} & \textbf{69.7} \\
\bottomrule
\end{tabular}
}
\end{center}
\vspace{-15pt}
\end{table}

\begin{table}[h]
\caption{Top-1 accuracy on ImageNet-LT with ResNet-50. Combination of \name~and state-of-the-art ensemble-based method RIDE \cite{wang2020long} can further improve its performance.}
\vspace{-5mm}
\label{tab:ride}
\begin{center}
\resizebox{.45\textwidth}{!}{
\begin{tabular}{c|cccc}
\toprule
Methods  & Many & Medium & Few & All \\ 
\midrule
RIDE (2 experts) \cite{wang2020long} & 65.8  & 51.0 & 34.6 & 54.4  \\
RIDE (3 experts) \cite{wang2020long} & 66.2 & 51.7 & 34.9 & 54.9 \\
RIDE (4 experts) \cite{wang2020long} & 66.2 & 52.3 & 36.5 & 55.4 \\
\midrule
\name+RIDE (2 experts) & 68.4 & 51.3 & 36.4 & 55.9 \\
\name+RIDE (3 experts) & 69.1 & 51.7 & 36.7 & 56.3 \\
\name+RIDE (4 experts) & \textbf{69.2} & \textbf{52.4} & \textbf{37.9} & \textbf{56.9} \\
\bottomrule
\end{tabular}
}
\end{center}
\vspace{-20pt}
\end{table}

\textbf{Combination of \name~and ensembling method}: In previous results, we compare \name~with established state-of-the-art single model baselines. Here we show that \name~can also be combined with a state-of-the-art ensemble-based method, RIDE \cite{wang2020long}, to further boost its performance. To implement \name~with RIDE, we simply replace the original stage-1 training in RIDE with \name~and keep the stage-2 routing training unchanged. As shown in Table \ref{tab:ride}, the combination of \name~and RIDE observes consistent improvements across all different number of experts. This further demonstrate the effectiveness of \name~on long-tailed recognition tasks.


\section{Analysis}
\label{sec:analysis}
We conduct an extensive analysis of \name~to explain its advantages over the baselines. We also conduct thorough ablations to demonstrate the effectiveness of each component in the \name~pipeline. 

\subsection{Understanding the Learned Representations}
\label{subsec:metrics}
In contrastive learning the features are regularized to fall on a hypershpere~\cite{chen2020simple,he2020momentum}, and the loss directly optimizes the distance between instances in the feature space. Thus, we can use distance in the feature space to evaluate the quality of learned representations. We propose several metrics to evaluate representations learned from long-tailed dataset, and study why \name~achieves better performance on long-tailed recognition than past work, e.g., KCL.

\textbf{Intra-Class Alignment.} One optimization goal of the contrastive loss is to minimize the distance between positive samples. Similar to \cite{wang2020understanding}, we define alignment under the supervised contrastive learning setting as the average distances between samples from the same class, where $F_i$ is the set of features from class $i$:
\begin{equation}
    \textbf{A}=\frac{1}{C}\sum_{i=1}^C \frac{1}{|F_i|^2}\sum_{v_j, v_k\in F_i}||v_j-v_k||_2.
\end{equation}

\textbf{Inter-Class Uniformity.} Another optimization goal of contrastive loss is to maximize the distance between negative samples. Hence, we define inter-class uniformity under the supervised contrastive learning setting as the average distances between different class centers:
\begin{equation}
    \textbf{U}=\frac{1}{C(C-1)}\sum_{i=1}^C\sum_{j=1, j\neq i}^C||c_i-c_j||_2,
\end{equation}
where $c_i$ is the center of samples from class $i$ on the hypersphere: $c_i = \frac{\sum\limits_{v_j \in F_i} v_j}{||\sum\limits_{v_j \in F_i} v_j||_2}$.

\begin{figure}[t]
\begin{center}
\includegraphics[width=0.4\textwidth]{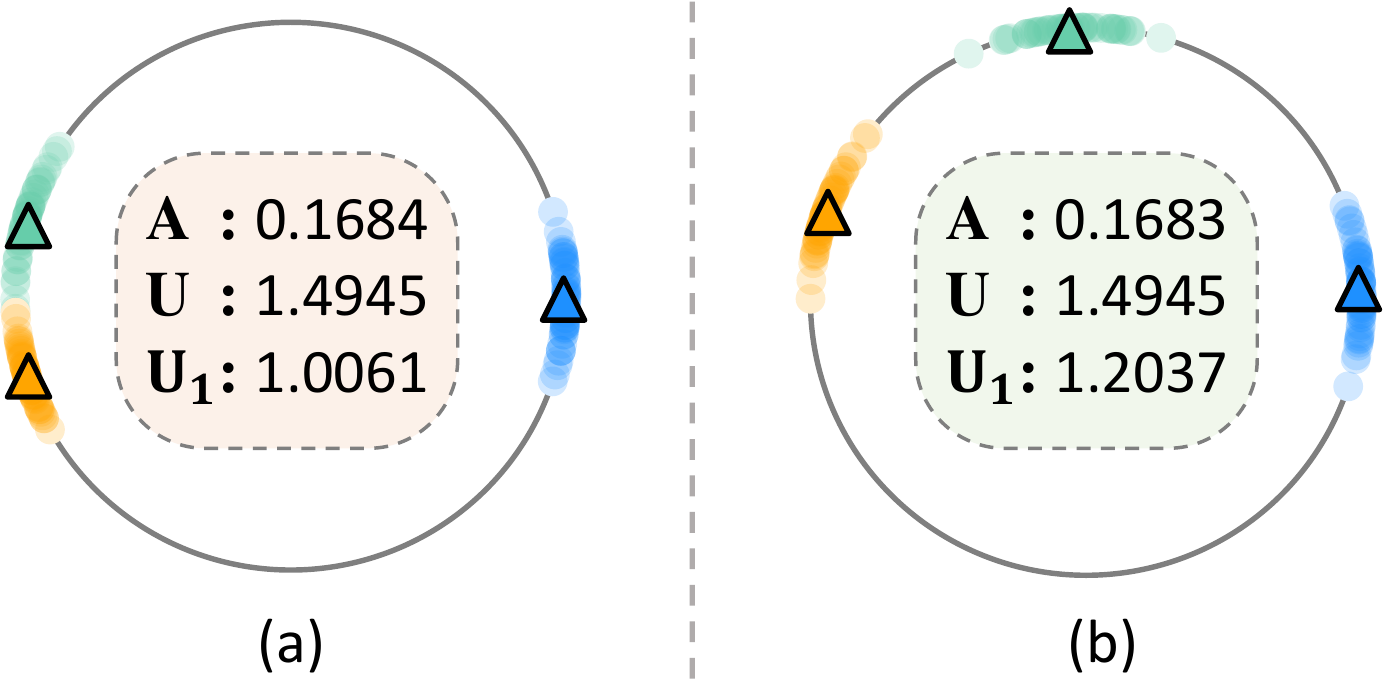}
\end{center}
\vspace{-5pt}
\caption{\small
Example feature distributions of 3 classes with the same intra-class alignment and inter-class uniformity but different nearest neighborhood uniformity $U_1$. Because the nearest neighbor class is too close, the decision boundary between the green class and the orange class in (a) is not clean.}
\label{fig:balance}
\vspace{-10pt}
\end{figure}

\textbf{Neighborhood Uniformity.} Though intra-class alignment and inter-class  uniformity are important metrics to measure the quality of learned representations, they cannot evaluate how close one class is to its neighbors. For example, as shown in Fig. \ref{fig:balance}, although both (a) and (b) achieve the same alignment and uniformity, (b) shows better neighborhood uniformity in the feature space and clearer decision boundaries between the green class and the orange class. 

Since what we really care about is only those classes that are too close to each other because the decision boundaries between them can be unclear, we define neighborhood uniformity as the distance to the top-$k$ closest class centers of each class:
\begin{equation}
    \textbf{U}_k=\frac{1}{Ck}\sum_{i=1}^C\min_{j_1,\cdots, j_k}(\sum_{l=1}^k||c_i-c_{j_l}||_2),
\end{equation}
where $j_1, \cdots, j_k \neq i$ are different classes.

\begin{figure}[h]
\begin{center}
\includegraphics[width=0.48\textwidth]{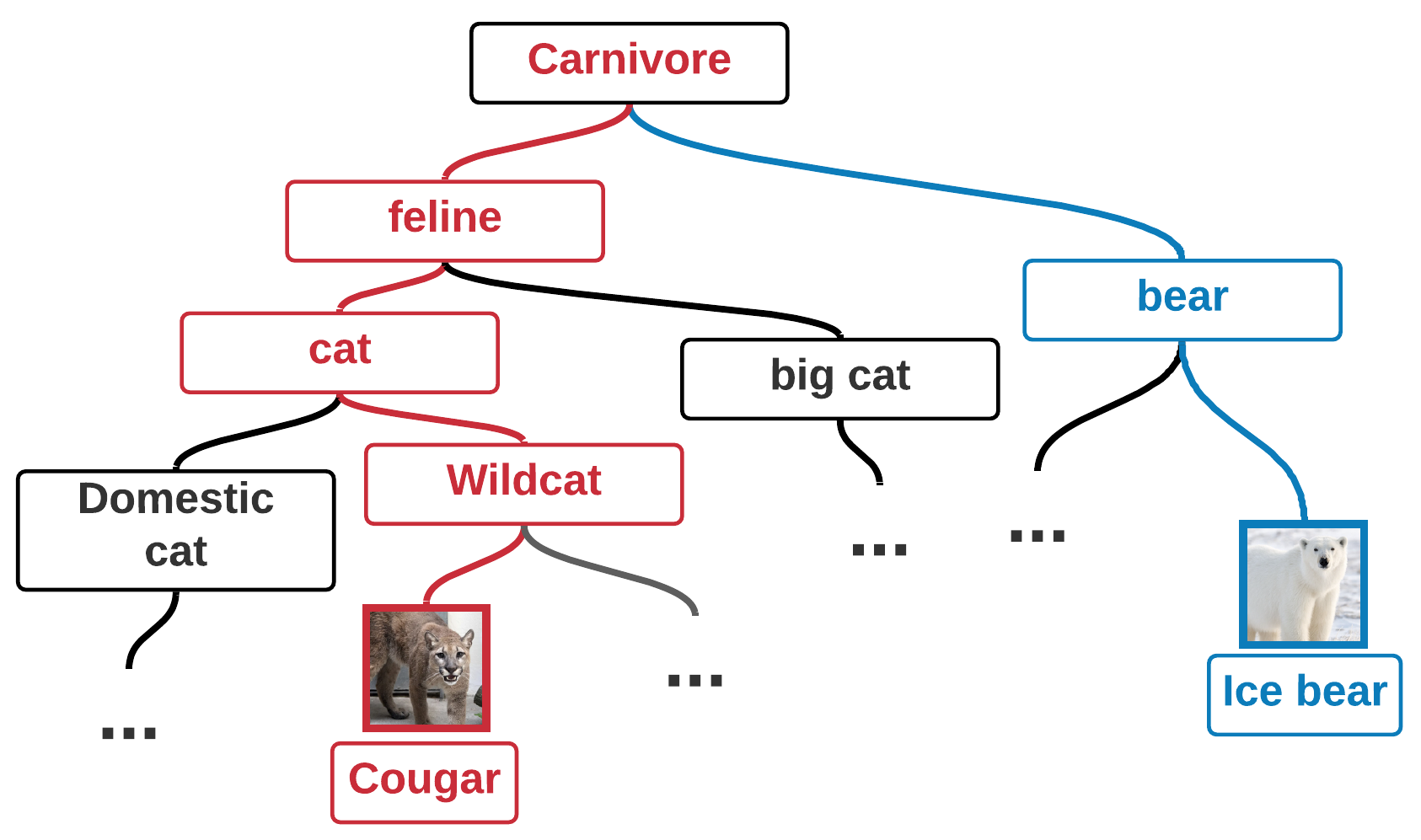}
\end{center}
\vspace{-15pt}
\caption{Illustration of semantic distance between two classes.}
\label{fig:semantic}
\vspace{-5pt}
\end{figure}

\textbf{Reasonability.} For better generalization, the learned feature space should also keep a reasonable semantic structure, i.e. ,classes that are semantically close to each other should also be close in feature space. Therefore, we define reasonability as the semantic distance between each class and its top-$k$ closest classes. The semantic distance of two classes is computed using WordNet hierarchy \cite{miller1995wordnet}, which is a hierarchical structure that contains all ImageNet classes as leaf nodes. The semantic distance of two classes is then defined as the shortest distance between the two leaf nodes in WordNet hierarchy. For example, as shown in Fig. \ref{fig:semantic}, the semantic distance between Cougar and Ice bear is 6.

\begin{table}[h]
\caption{\name~achieves better uniformity, neighborhood uniformity and reasonability than KCL on ImageNet-LT, while keeping almost the same alignment. The $k$ for neighborhood uniformity and reasonability is set to 10. $\uparrow$ indicates larger is better, whereas $\downarrow$ indicates smaller is better.}
\vspace{-5mm}
\label{tab:metrics}
\begin{center}
\resizebox{.45\textwidth}{!}{
\begin{tabular}{cc|cccc}
\toprule
Metric & Methods  & Many & Medium & Few & All \\ 
\hline
\multirow{2}{*}{\textbf{A}$^\downarrow$} & KCL$\dagger$ & 0.71 & \textbf{0.69} & \textbf{0.72} & \textbf{0.70}\\
& \name & 0.71 & 0.70 & 0.74 &  0.71\\
\hline
\multirow{2}{*}{\textbf{U}$^\uparrow$} & KCL$\dagger$ & 1.33 & 1.32 & 1.30 & 1.32\\
& \name & \textbf{1.38} & \textbf{1.38} & \textbf{1.37} &  \textbf{1.38}\\
\hline
\multirow{2}{*}{\textbf{U}$_{10}^\uparrow$} & KCL$\dagger$ & 0.94 & 0.89 & 0.87 &  0.91\\
& \name & \textbf{1.02} & \textbf{1.02} & \textbf{1.05} & \textbf{1.02} \\
\hline
\multirow{2}{*}{\textbf{R}$^\downarrow$} & KCL$\dagger$ & 7.35 & 7.25 & 7.42 &  7.31\\
& \name & \textbf{7.22} & \textbf{7.13} & \textbf{6.94} &  \textbf{7.14}\\
\hline
\multirow{2}{*}{Acc.$^\uparrow$ } & KCL$\dagger$ & 62.4 & 49.0 & 29.5 & 51.5\\
& \name & \textbf{63.5} & \textbf{49.7} & \textbf{30.4} &  \textbf{52.4}\\
\bottomrule
\end{tabular}}
\end{center}
\vspace{-15pt}
\end{table}

\textbf{KCL vs. \name.} In Table \ref{tab:metrics}, we compare the alignment, uniformity, neighborhood uniformity and reasonability of KCL and \name~on ImageNet-LT. The results highlight several good properties of \name~over KCL: 1.) \name~achieves better uniformity, neighborhood uniformity, and reasonability than KCL on all class splits, while keeping almost the same alignment as KCL. 2.) Although \name's uniformity is only 0.06 higher than that of KCL, its neighborhood uniformity (the average uniformity of the closest 10 classes) is 0.17 higher than that of KCL. Moreover, KCL's neighborhood uniformity is even worse on tail classes, while \name~keeps consistent neighborhood uniformity over all classes. This demonstrates the effectiveness of \name~on keeping all classes away from each other, and thus allowing clearer decision boundaries between classes. 3.) \name~achieves better reasonability than KCL, especially on tail classes, showing that the learned feature space is not only uniform but also semantically reasonable.


\begin{figure}[h]
\begin{center}
\includegraphics[width=0.46\textwidth]{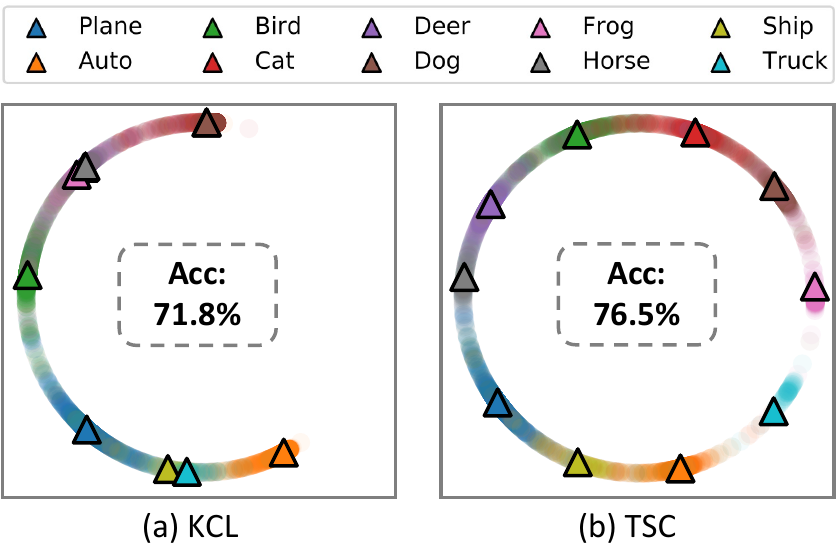}
\end{center}
\vspace{-7.5pt}
\caption{Feature distribution and accuracy of (a) KCL and (b) \name~on CIFAR-10-LT with imbalance ratio 100. In (a), the class centers of cat and dog collapse together, as well as the class centers of horse, deer, and frog.}
\vspace{-7.5pt}
\label{fig:viz}
\end{figure}

\textbf{Visualization.} We can obtain more insights by visualizing the features learned by \name\ and KCL. 
In Fig.~\ref{fig:viz}, we visualize the features learned using KCL and \name~on CIFAR-10-LT with imbalance ratio of 100 and $d=2$. The triangles indicate the class centers. As shown in Fig.~\ref{fig:viz}(a), features learned by KCL suffer from poor uniformity. Several class pairs collapse into each other, leading to unclear  boundaries. We also see that when uniformity is quite poor, a large part of the feature space is left empty. On the other hand, Fig. \ref{fig:viz}(b) shows that features learned using \name\ achieve good uniformity and clear separation between classes, hence achieve better classification performance.

\subsection{Ablations}
\label{subsec:ablation}

\begin{table}[h]
\caption{\small Comparison between KCL, KCL with class-balanced sampling and \name~on CIFAR-10-LT and CIFAR-100-LT.}
\vspace{-5mm}
\label{tab:cb}
\small
\begin{center}
\resizebox{.48\textwidth}{!}{
\begin{tabular}{c|c|c|c|c|c|c}
\toprule
Dataset    &   \multicolumn{3}{c|}{CIFAR-10-LT}    &    \multicolumn{3}{c}{CIFAR-100-LT}   \\
\midrule
Imbalance Ratio ($\rho$)  & 100    &    50    &    10    &   100    &   50   &   10   \\
\midrule\midrule
KCL$\dagger$ & 77.6 & 81.7 & 88.0 & 42.8 & 46.3 & 57.6  \\
CB-KCL & 75.5 & 80.2 & 87.1 & 41.5 & 45.5 & 56.8 \\
\midrule
\name & \textbf{79.7} & \textbf{82.9} & \textbf{88.7}         & \textbf{43.8}  &  \textbf{47.4}  & \textbf{59.0} \\
\bottomrule
\end{tabular}
}
\end{center}
\vspace{-15pt}
\end{table}

\textbf{Class-Balanced Sampling.} Since KCL exhibits poor uniformity, one may think of improving its uniformity using class-balanced sampling. Table \ref{tab:cb} compares KCL with class-balanced sampling with \name. As shown in the table, class-balanced sampling gets even worse performance than standard KCL on both CIFAR-10 and CIFAR-100. This phenomenon is also shown in \cite{kang2019decoupling}, where the author shows that instance-balanced sampling achieves the best results among different sampling strategies during representation learning. \name\ uses instance-balanced sampling, and achieves good uniformity using pre-computed targets.

\begin{table}[h]
\caption{Balanced positive sampling strategy improves less on TSC than on FCL on ImageNet-LT.}
\vspace{-5mm}
\label{tab:kclvsfcl}
\begin{center}
\resizebox{.43\textwidth}{!}{
\begin{tabular}{c|cccc}
\toprule
Methods  & Many & Medium & Few & All \\ 
\midrule
FCL \cite{kang2020exploring} & 61.4 & 47.0 & 28.2 & 49.8 \\
KCL \cite{kang2020exploring} & 61.8 & 49.4 & 30.9 & 51.5 \\
\name~(FCL) & 62.7 & 49.2 & 30.1 & 51.8 \\
\name~(KCL) & 63.5 & 49.7 & 30.4 & 52.4 \\
\bottomrule
\end{tabular}
}
\end{center}
\vspace{-15pt}
\end{table}

\textbf{Benefits of Balanced Positive Samples.} \cite{kang2020exploring} has shown that sampling positive pairs in a balanced way (as done in KCL) is better than taking all samples of the same class as positives (as done in FCL). Therefore, \name~also builds on top of the KCL loss which samples the same number of positive pairs for each data point. However, we also notice that this balanced positive sampling strategy brings much less improvement to \name~than KCL. As shown in Table \ref{tab:kclvsfcl}, the balanced positive sampling strategy in KCL improves 1.7\% over FCL, while improves 0.7\% over \name~with FCL. This is possibly because with a balanced feature space, the alignment within each class is also naturally balanced and therefore does not need the balanced positive sampling strategy, which further demonstrates the importance of a balanced feature space.

\begin{figure}[h]
\begin{center}
\includegraphics[width=0.46\textwidth]{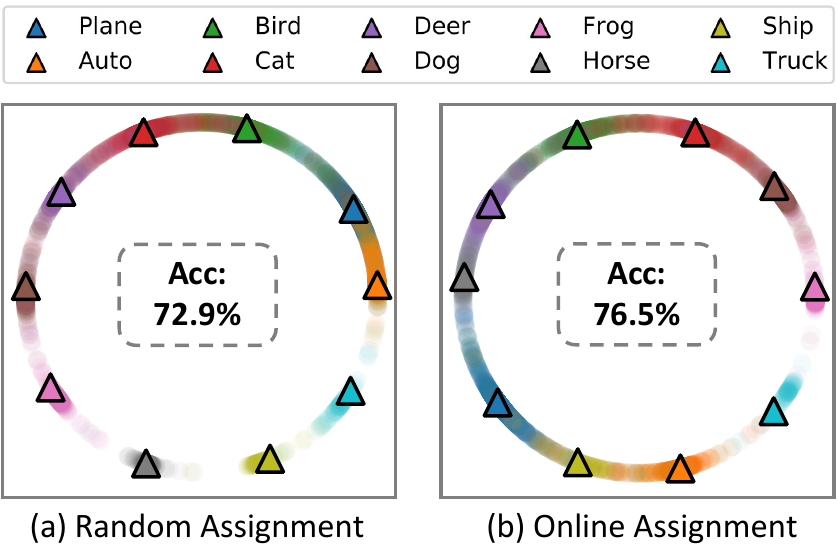}
\end{center}
\vspace{-10pt}
\caption{
Comparison of feature distribution and accuracy using (a) random class assignment and (b) online assignment in \name~ on CIFAR-10-LT with imbalance ratio 100.}
\label{fig:matching}
\vspace{-10pt}
\end{figure}

\begin{table}[h]
\caption{\name~with and without online matching algorithm on ImageNet-LT.}
\vspace{-5mm}
\label{tab:matching}
\begin{center}
\resizebox{.45\textwidth}{!}{
\begin{tabular}{c|ccccc}
\toprule
Methods  & Many & Medium & Few & All &  \textbf{R}$^\downarrow$\\ 
\midrule 
KCL $\dagger$ & 62.4 & 49.0 & 29.5 & 51.5 & 7.31 \\
\name~(random assign) & 61.8 & 48.1 & 29.2 & 50.8 & 7.81\\
\name~(online matching) & \textbf{63.5} & \textbf{49.7} & \textbf{30.4} & \textbf{52.4} & \textbf{7.14}\\
\bottomrule
\end{tabular}
}
\end{center}
\vspace{-15pt}
\end{table}

\textbf{Online Matching Algorithm.} In Fig. \ref{fig:matching}(a), we show the performance of \name~with randomly assigned targets on CIFAR-10-LT, where the target for each class is randomly assigned at the beginning of training and fixed for the entire training process. To better visualize the features, the feature dimension of the output is set to 2. As shown in the figure, both methods achieve good uniformity on training data. However, the semantics in Fig. \ref{fig:matching}(a) are not reasonable, as semantically close classes are not nearby in the feature space, e.g., deer and horse. Similar results are also shown in Table \ref{tab:matching}, where we compare the reasonability of \name~with and without the online matching algorithm. Without the online matching algorithm, the reasonability of \name~is significantly worse than with it, resulting in much poorer generalization performance.

\begin{table}[h]
\caption{$\mathcal{L}_u$ achieved with different random seeds during optimal targets generation for different numbers of classes.}
\vspace{-5mm}
\label{tab:targets}
\begin{center}
\resizebox{.35\textwidth}{!}{
\begin{tabular}{c|cccc}
\toprule
\# Class  & 10 & 100 & 1000 & 8142 \\ 
\midrule 
seed = 0 & 14.286 & 14.286 & 14.287 &14.297\\
seed = 1 & 14.286 & 14.286& 14.287 &14.297\\
seed = 2 & 14.286 & 14.286& 14.287 &14.297\\
seed = 3 & 14.286 & 14.286& 14.287 &14.297\\
seed = 4 & 14.286 & 14.286& 14.287 &14.297\\
\midrule
std &3.2e-6  & 6.1e-6& 1.8e-6& 3.0e-6\\
\bottomrule
\end{tabular}
}
\end{center}
\vspace{-10pt}
\end{table}

\textbf{Stability of Target Generation.} An important step of our pipeline is to generate optimal targets. Since we use numerical approximation (SGD) to generate optimal targets, it is possible that the generated targets can achieve different minimum of $\mathcal{L}_u$ with different random seeds. Here we show the stability of our targets generation process. Table \ref{tab:targets} shows the final $\mathcal{L}_u$ achieved by SGD with different random seeds. As shown in the table, with different random seed, the final $\mathcal{L}_u$ stays quite stable with negligible standard deviation. Therefore, the optimal targets generation process is stable.

\section{Conclusion \& Limitations}
In this paper, we introduced targeted supervised contrastive learning (\name) for long-tailed recognition. We empirically showed that, for unbalanced data, features learned by traditional supervised contrastive losses lead to reduced uniformity and unclear class boundaries, and hence poorer performance. By assigning uniformly distributed targets to each class during training, \name~avoids this problem, leading to a more uniform and balanced feature space. Extensive experiments on multiple datasets show that \name\ achieves state-of-the-art single-model performance on all benchmark datasets for long-tailed recognition.

Nonetheless, \name\ has some limitations. First, the optimal targets of \name\ are computed using stochastic gradient descent. The analytical optimal solution of points on a hypersphere that minimize an energy potential remains an open problem (Thomson problem). Though \name\ uses an approximate solution, the empirical results show consistent and significant performance gain. Second, \name~requires knowing the number of classes in advance to compute the targets; so, it is not applicable to problems where the number of classes is unknown. Despite these limitations, we believe that \name~provides an important step forward for long-tailed recognition, and delivers new insights on how data distribution affects key properties of contrastive learning.



 {\small
 \bibliographystyle{ieee_fullname}
 \bibliography{main}
 }





\end{document}